\documentclass[journal,letter]{IEEEtran}

\usepackage{cite}
\usepackage[pdftex]{graphicx}
\DeclareGraphicsExtensions{.pdf}
\usepackage{array} % neatens table row-spacing
\usepackage[caption=false,font=footnotesize]{subfig}
\usepackage{booktabs} % professional tables
\usepackage{amsmath}
\usepackage{gensymb} % for the \degree symbol
\usepackage{color} % to highlight review suggestions

\begin{document}

\title{View-invariant Gait Recognition through \\ Genetic Template Segmentation}

\author{Ebenezer~R.H.P.~Isaac~\IEEEmembership{Member,~IEEE,} Susan~Elias,
  Srinivasan Rajagopalan, and~K.S.~Easwarakumar% <-this % stops a space
  \thanks{Accepted manuscript (June 10, 2017). Please refer the published version in http://dx.doi.org/10.1109/LSP.2017.2715179. Vol. 24, No. 8, Aug. 2017.}% 
  \thanks{E.R.H.P. Isaac and K.S. Easwarakumar are with the Department of
    Computer Science and Engineering, Anna University, Chennai, India
    (e-mail: ebeisaac@ieee.org, easwarakumarks@gmail.com)}% <-this % stops a space
  \thanks{Susan Elias is with the School of Electronics Engineering, VIT
    University, Chennai Campus, India (e-mail:
    susan.elias@vit.ac.in)}% <-this % stops a space
  \thanks{Srinivasan Rajagopalan is with Department of Physiology and Biomedical
    Engineering, Mayo Clinic, Rochester, Minnesota, USA (email:
    rajagopalan.srinivasan@mayo.edu)}}% <-this % stops a space

%The paper headers
\ifCLASSOPTIONpeerreview
\markboth{IEEE Signal Processing Letters,~Vol.~XX, No.~X, Month~20XX}%
{E. Isaac \MakeLowercase{\textit{et al.}}: Gait Recognition through Genetic
  Template Segmentation}
\else
\markboth{IEEE Signal Processing Letters,~Vol.~XX, No.~X, Month~20XX}%
{Gait Recognition through Genetic Template Segmentation}
\fi
\maketitle

\begin{abstract}  
  Template-based model-free approach provides by far the most successful
  solution to the gait recognition problem in literature.  Recent work discusses
  how isolating the head and leg portion of the template increase the
  performance of a gait recognition system making it robust against covariates
  like clothing and carrying conditions. However, most involve a manual
  definition of the boundaries. The method we propose, the genetic template
  segmentation (GTS), employs the genetic algorithm to automate the boundary
  selection process. This method was tested on the GEI, GEnI and AEI
  templates. GEI seems to exhibit the best result when segmented with our
  approach.  Experimental results depict that our approach significantly
  outperforms the existing implementations of view-invariant gait recognition.
\end{abstract}


\ifCLASSOPTIONpeerreview
\begin{center} \bfseries EDICS Category: MLSAS-PATT, IMD-PATT\end{center}
\fi

\IEEEpeerreviewmaketitle

\section{Introduction}

\IEEEPARstart{G}{ait} recognition analyses the manner of walking for human
identification. As it requires minimal cooperation from the subject compared to
other modalities, it is considered to be an unobtrusive biometric. Gait
recognition methods can be grouped into to either model-based or model-free
approaches. Model-based methods \cite{bouchrika2007model, Goffredo2008, Zhang_R,
  yam2004automated} attempt to track the dynamic changes in the articulation
points during gait and hence require intense computational effort. Recent trends prefer
the model-free approach as it captures the gait patterns without this requirement. 

The notion of templates was introduced in \cite{collins2002silhouette} where the
silhouettes of key frames are matched with that of the gallery
for recognition. Han and Bhanu \cite{man2006individual} projected a simple
method that averages all silhouettes of a single gait cycle to produce a single
image template called the gait energy image (GEI) to encompass the
spatiotemporal characteristics. Its advent brought forth a new category of
model-free gait recognition called template-based methods. The GEI quickly
became the most successful method for multi-view gait recognition. Its major
drawback was its weakness to covariates like clothing and load carrying which
could adversely affect its performance. Many similar methods followed aiming to
mitigate this weakness with their implementation of gait templates. Two of such
notable templates were the active energy image (AEI)~\cite{zhang2010active} and
the gait entropy image (GEnI)~\cite{bashir2010gait}. With a slight trade-off in
normal walk gait recognition, these new templates were able to produce a better
recognition accuracy over the clothing and carrying covariates in gait. Bashir
\textit{et al.}  \cite{bashir2010gait} eliminated this trade-off by masking the
GEI with the image of the respective GEnI.

In addition to clothing and carrying conditions, the view angle is found to be the
most important covariate factor that affects gait recognition performance
\cite{zeng2016view, huang2008human, cilla2012probabilistic, liu2013robust}.
There are essentially two types of view-invariant gait recognition models: view
transformation model (VTM) and view-preserving model (VPM).

VTMs \cite{kusakunniran2010support, kusakunniran2012gait, Zhao2015} transform
the probe sequence's angle to match with that of the gallery sequence. The VTM
methods may differ in the measures used to gauge the transformation accuracy
\cite{muramatsu2016view}. However, a significant level of error is inevitable in
VTM-based gait recognition \cite{nini2011, dupuis2013feature}.

VPMs consider multiple views as part of the gallery itself. This process
incorporates the view information within the feature set for the extraction of
relevant view-invariant gait features. Various methods can be employed to
facilitate this. Examples include varying width vectors \cite{zeng2016view},
Grassmann manifold \cite{connie2016grassmannian}, geometric view estimation
\cite{jia2015view}, and spatiotemporal feet positioning
\cite{verlekar2016view}. A variant of VPM involves
extraction of view-independent features through multi-view training and then use
a single gallery view for testing \cite{nini2011,tang2017robust}. 

Dupuis \textit{et al.} \cite{dupuis2013feature} formulated a single mask through
the ranking of pixel features using the Random Forests classifier. Their
panoramic gait recognition (PGR) algorithm uses pose estimation for view
prediction. Choudhury \textit{et al.} \cite{choudhury2015robust} designed a VPM
named view-invariant multiscale gait recognition (VI-MGR) which applied
Shannon's entropy function to the lower limb region of the GEI. The sub-region
selection was later modified by Rida \textit{et al.}  \cite{rida2016human}
automating this segmentation procedure with a process known as group lasso of
motion (GLM). Their approach to the problem has shown significant improvement in
the covariate recognition accuracy.

Though the following implementations do not concern view-invariance or covariate
factors, their aspects add to the motivation of our approach.  Jia \textit{et
  al.} \cite{jia2015view} have shown how incorporating the head and shoulder mean
shape (HSMS) along with the Lucas-Kanade variant of the gait flow image (GFI)
\cite{lam2011gait} greatly improves recognition accuracy. The genetic algorithm
\cite{Goldberg} was previously used in \cite{yeoh2014genetic} to optimize the
selection of model-based gait parameters and also in \cite{tafazzoli2015genetic}
for the selection of superimposed contour features.

\begin{figure}
  \centering
  \includegraphics[width=0.93\linewidth]{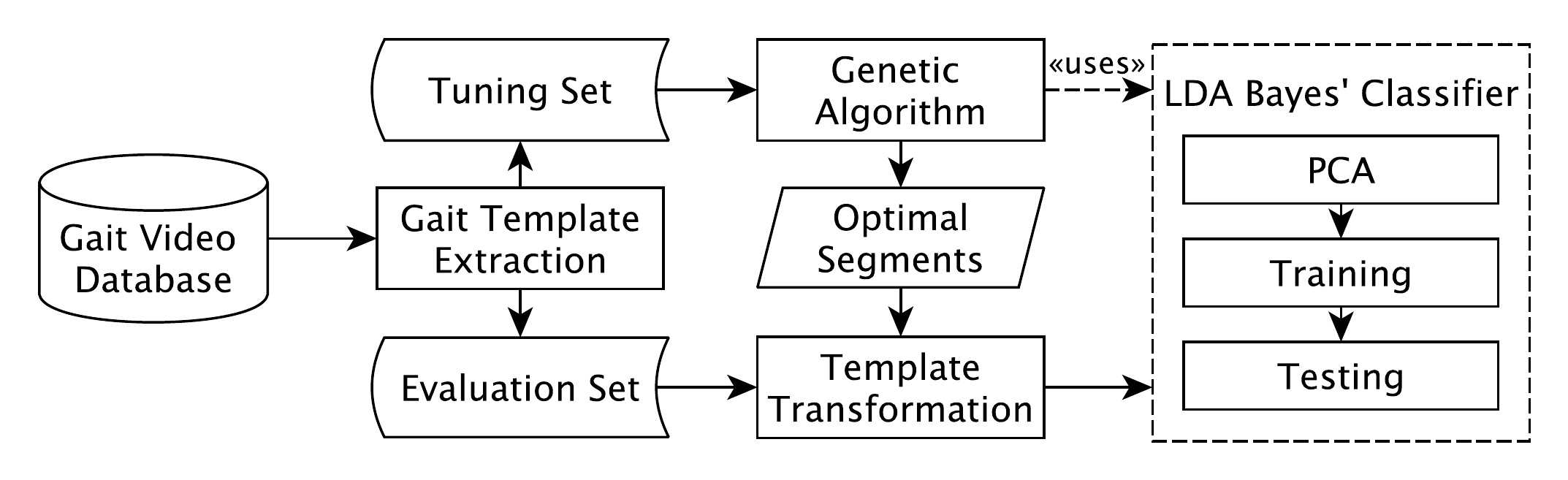}
  \caption{A simplified flow of the proposed method.}
  \label{fig:arch}
\end{figure}

In this article, we devise a VPM that can be applied to any gait template
for gait identification. To refine the templates themselves, a method is
proposed to automate its segmentation process with the use of the genetic
algorithm (GA). These segments depict the optimal regions of the gait template
that can be used to obtain the best recognition result at any covariate
factor. The contributions of this paper are summarized as follows:

\begin{itemize}
\item A sub-region selection process through GA that greatly enhances the
  robustness of gait recognition against covariate factors.
\item A separate mask is produced for multiple view angles to obtain the
  best possible feature set for any given view.
\item A computationally efficient view-estimator design to detect the angle of
  view based on the slopes of the gait trajectory.
\end{itemize}

\section{Method}
\label{sec:method}

An overview of the method is illustrated in Fig.~\ref{fig:arch}. The first step
is to extract the gait template (such as the GEI) from the video that contains
the gait sequence. After which the database is split into two disjoint sets --
tuning set and evaluation set. The tuning set is fed to the GA to formulate the
segments for optimal performance. Only those segments are extracted from the
evaluation set to test the final accuracy of the system. The features are
preprocessed by Principal Component Analysis (PCA) followed by a multi-class
Linear Discriminant Analysis (LDA) and then classified using Bayes' rule. The
Multi-class LDA, also referred to as Multiple Discriminant Analysis
(MDA)~\cite{duda2001pattern}, is a supervised dimensionality reduction method
that would maximize inter-class distance while minimizing intra-class distance.
PCA~\cite{jolliffe2002PCA} is an unsupervised dimensionality reduction algorithm
that projects the given features to feature space that corresponds to the
highest variance. The use of PCA yeilds a net positive effect on the performance
of the classifier in terms of both processing time and accuracy. As a design choice, we use Bayes' rule over the widely adopted $k$NN.

%As $k$NN is a lazy learner, it requires the whole of the processed training set to reside in the system while Bayes' rule needs only to store its learned model parameters.

\subsection{Gait Template Extraction}
\label{sec:gait-templ-extr}

All gait templates are produced in a similar procedure to the one given
below. Silhouettes in here are obtained through background subtraction and
encoded in grayscale.

\begin{enumerate}
\item Extract only the silhouettes of the subject during a single gait cycle.
\item The silhouettes are center-aligned and scaled to a standard size; 240 x
  240 in this case.
\item The standardized silhouettes for a given gait sequence are merged through
  a collation process to generate the gait template.
\end{enumerate}

Let $N$ be the number of silhouettes for a gait cycle for a given subject. Each
$t^\text{th}$ silhouette is denoted as $B(t)$. The novelty in a gait template is
defined by its collation process. For example, in GEI \cite{man2006individual},
the collation process is given by
\[G_\text{GEI} = \frac{1}{N}\sum_{t=1}^{N}B(t)\]
Similarly, the templates AEI~\cite{zhang2010active} and GEnI~\cite{bashir2010gait} used in this study also differ by their collation process.

\subsection{Genetic Template Segmentation}
\label{sec:chromosome}

\begin{figure}
  \centering
  \includegraphics[width=0.3\linewidth]{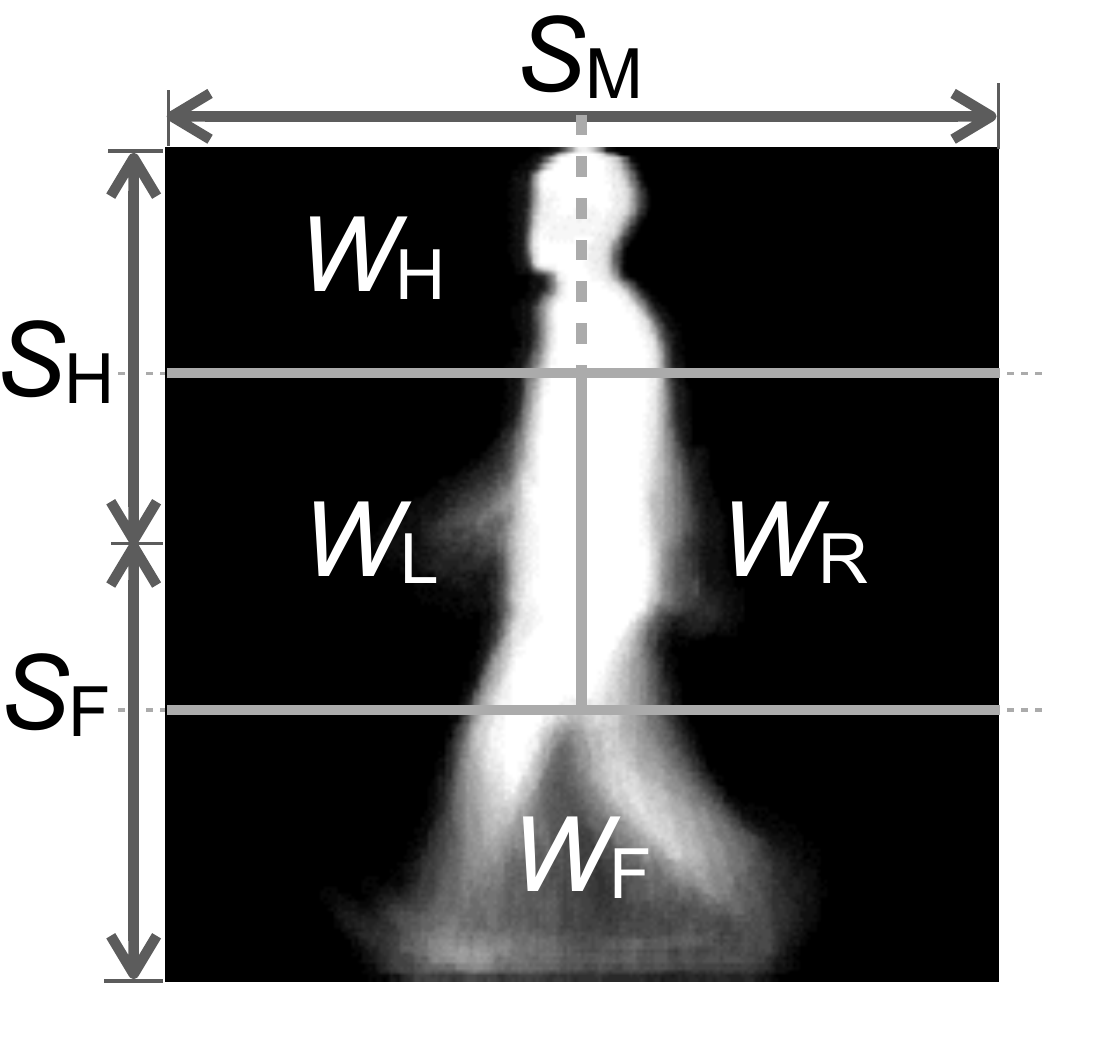}
  \caption{Parameters optimized by the genetic algorithm. Note that the split
    variables $S_i$ are scalars while the weight variables $W_i$ are bits.}
  \label{fig:genparams}
\end{figure}

The boundary selection process is automated through GA to find
the optimum boundary to segment the gait template before the actual training
process. The gait template is to be split into four segments, viz., head
portion H, leg portion F, mid-left section L and mid-right section R. The parameters to
be optimized are the split points to divide these sections and a binary
weight bit per region to decide whether the respective region should be included
in the training as shown in Fig.~\ref{fig:genparams}. This process is used to
produce a masking template for each view angle.

The chromosome structure for the genetic optimization is given as
\[[S_\text{H}, S_\text{M}, S_\text{F}, W_\text{H}, W_\text{L}, W_\text{R}, W_\text{F}]\]

The variables denoted $S_i$ are split variables that determine the boundary for the
region to segment and is represented by 8 bits each. $S_\text{H}$ defines the
line between the head portion and the midsections; $S_\text{F}$ determines
the split between the midsections and leg region; $S_\text{M}$ divides the two
midsections. If $d$ is the decimal equivalent of the 8 bits used to represent
the split variables, then its value can be decoded as
\[S_i = \text{min}_i + (\text{max}_i-\text{min}_i) \times d_i/255\]

where min$_i$ and max$_i$ are the minimum and maximum possible values for the
variable $S_i$. The variables $W_i$ are binary variables that determine
whether the segment is included for training, 1 indicates inclusion while 0
represents masking. The total size of chromosome hence becomes 28 bits.

A set of subjects with all covariates included is used as a tuning set to
determine boundary locations for segmentation. The fitness function evaluates
the hypothesis generated by the chromosome against the tuning set to produce a
fitness measure. The three covariates considered here is A: normal walk, B:
carrying a bag and C: clothing condition. If the fitness measure is simply set
to the average of the accuracy of the three covariate sets, then the GA would
make a significant trade-off on the normal walk sequence to maximize the overall
accuracy. This was experimentally observed to at 90\% while the state-of-the-art
approaches produce accuracies of above 95\% \cite{rida2016human}. The fitness
measure, $F$ for a given chromosome, $h$, is calculated as
\[F(h) = \Big(\frac{1}{2}\text{CCR}_\text{A}(h) 
  + \frac{1}{6}\text{CCR}_\text{B}(h) 
  + \frac{1}{3}\text{CCR}_\text{C}(h)\Big)^2 \]

where CCR$_K$ represents the correct classification rates for the corresponding
covariate $K$. Giving equal weights to each of the CCR$_k$ causes a trade-off in
normal condition performance leading to an accuracy of 95.6\% which is among the
lowest of the normal CCR (refer Table~\ref{tab:ccr}). Thus, the highest priority
was given to CCR of the normal setting, CCR$_\text{A}$, to compete with the
state of the art. In most approaches, clothing conditions pose the greatest
challenge to template-based recognition systems. Hence the accuracy pertaining
to the clothing condition, CCR$_\text{C}$, was given the next highest weight
after normal setting to boost its accuracy on par with the carrying
condition, CCR$_\text{B}$. These priority weights were assigned empirically. 

The elitist selection variant of the generation propagation is used for this
implementation of the GA \cite{baluja1995removing}. That is, the
chromosome corresponding to the highest fitness of a generation T$_n$ is made sure to
be propagated the next generation T$_{n+1}$. The GA is set to follow
a uniform crossover with probability 0.6, a single bit mutation probability
of 0.03 and populates 20 chromosomes per generation. The optimization runs for
15 generations although convergence was mostly attained before the
8\textsuperscript{th} generation during experimental observation.

% \begin{figure}
%   % Recompile segmented images
%   \centering
%   \subfloat[]{\includegraphics[width=0.15\linewidth]{Org-GEI-090.png}%
%     \label{fig:org-gei}}
%   \hfil
%   \subfloat[]{\includegraphics[width=0.15\linewidth]{Seg-GEI-090.png}%
%     \label{fig:seg-gei}}
%   \hfil
%   \subfloat[]{\includegraphics[width=0.15\linewidth]{Org-GEnI-090.png}%
%     \label{fig:org-gni}}  
%   \hfil
%   \subfloat[]{\includegraphics[width=0.15\linewidth]{Seg-GEnI-090.png}%
%     \label{fig:seg-gni}}  
%   \hfil
%   \subfloat[]{\includegraphics[width=0.15\linewidth]{Org-AEI-090.png}%
%     \label{fig:org-aei}}
%   \hfil
%   \subfloat[]{\includegraphics[width=0.15\linewidth]{Seg-AEI-090.png}%
%     \label{fig:seg-aei}}
%   \caption{Template transformation using the GTS hypothesis. Original templates
%     of (a) GEI, (c) GEnI and (e) AEI, and their respective masked templates (b),
%     (d) and (f) after the genetic template segmentation.}
%   \label{fig:gtsmask}
% \end{figure}

\begin{figure}
  \centering
  \includegraphics[width=0.85\linewidth]{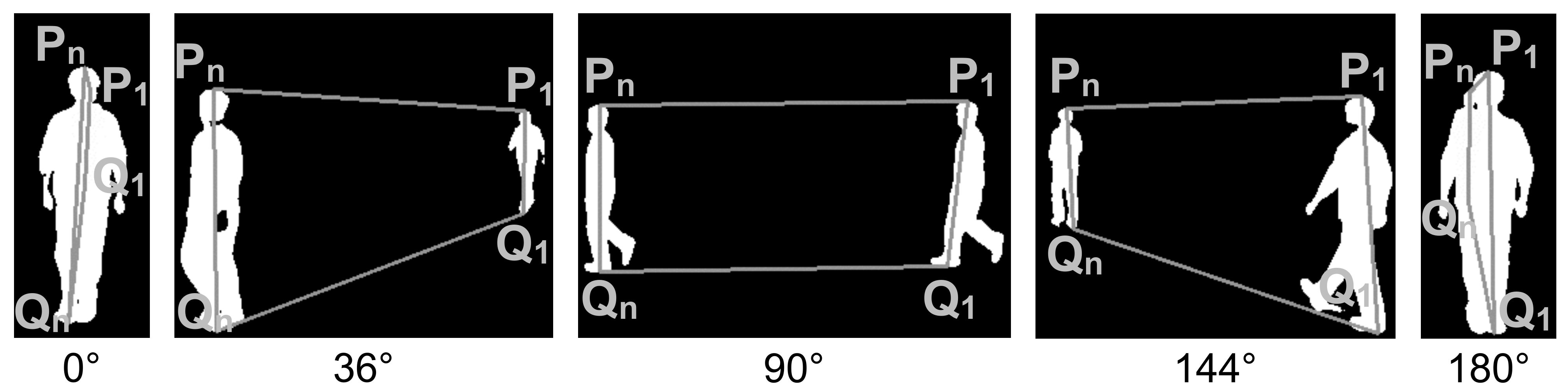}
  \caption{Estimating viewpoints from different views. The line joining the
    extremities depicts the path of gait.}
  \label{fig:viewest}
\end{figure}

\begin{figure}
  \centering
  \includegraphics[width=0.85\linewidth]{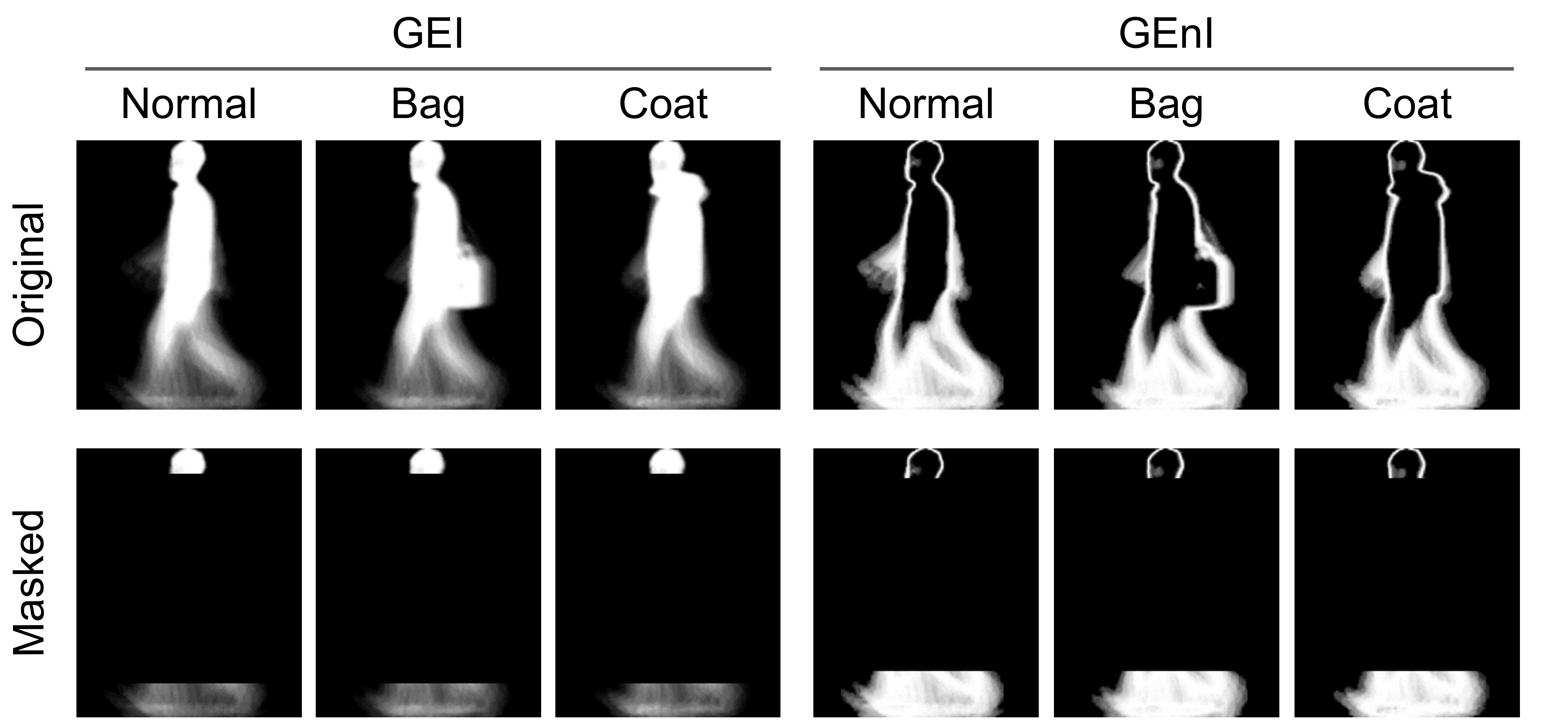}
  \caption{Template transformation using the GTS hypothesis. Original templates
    of the GEI and GEnI, and their respective masked templates for each
    covariate.}
  \label{fig:gtsmask}
\end{figure}

\subsection{View Estimation}
\label{sec:view-est}
Under the assumption that the subjects walk in a straight line for verification,
the first and last visible silhouettes, $S_1$ and $S_n$, are taken into
consideration. Let $P_1$ and $Q_1$ be the topmost and bottom-most point of $S_1$
as illustrated in Fig.~\ref{fig:viewest}.  Similarly, $P_n$ and $Q_n$ denote the
topmost and bottom-most point of $S_n$. Let $m_P$ and $m_Q$ be the slopes of the
lines $P_1P_n$ and $Q_1Q_n$ respectively. These two slopes alone form the
features required to train the view-estimation classifier with the view as
output labels. To reduce the number of cases, the sequence is passed through a
simple check to verify whether the angle lies in the coronal plane ($0\degree$
or $180\degree$). If the last silhouette overlaps the first, then the viewpoint
is determined to be at $0\degree$ and the direct opposite for $180\degree$. If
both of these cases fail, then the angle should be one among those other than
the two in the coronal plane.

\section{Experimental Results and Discussion}
\label{sec:results}

\begin{table}
{
  \centering
  \caption{{\hspace{3em} CCR(\%) of Different Algorithms on\newline 
      CASIA Dataset B at $90\degree$ view}}
  \label{tab:ccr}
  % RECOMPILE THIS TABLE BEFORE SUBMISSION
  %\includegraphics[width=\linewidth]{CCRtable.pdf}
\resizebox{\linewidth}{!}{
  \begin{tabular}{c p{3cm} *{5}{c}}
    \toprule
    Year & Method & Normal & Bag & Coat & Mean & Std \\
    \cmidrule(lr){1-2}
    \cmidrule(lr){3-5}
    \cmidrule(lr){6-7}
    2006 & Han and Bhanu \cite{man2006individual} & 99.60 & 57.20 & 23.80 & 60.20 & 37.99 \\
    2010 & Bashir et al. \cite{bashir2010gait} & \textbf{100.0} & 78.30 & 44.00 & 74.10 & 28.24 \\
    2013 & Dupuis et al. \cite{dupuis2013feature} & 98.43 & 75.80 & 91.86 & 88.70 & 11.64 \\
    2014 & Kusakunniran \cite{kusakunniran2014attr} & 94.50 & 60.90 & 58.50 & 71.30 & 20.13 \\
    2015 & Arora et al. \cite{Arora2015} & 98.00 & 74.50 & 45.00 & 72.50 & 26.56\\
    2015 & Yogarajah et al. \cite{Yogarajah20153} & 97.60 & 89.90 & 63.70 & 83.73 & 17.77 \\
    2016 & Rida et al. \cite{rida2016human} & 98.39 & 75.89 & 91.96 & 88.75 & 11.59 \\
    - & GEI with GTS & 98.00 & \textbf{95.50} & \textbf{93.00} & \textbf{95.50} &  \textbf{2.50} \\
    - & GEnI with GTS & 97.00 & 95.00 & 91.00 & 94.33 &  3.06 \\
    - & AEI with GTS & 89.50 & 85.50 & 77.50 & 84.17 &  6.11 \\
  \bottomrule
  \end{tabular}%
}
}

%\nocite{kusakunniran2014attr, Arora2015, Yogarajah20153}
  % \vspace{0.5em}
  % In all of the above, Set-A corresponds to normal walk where A1 is used as the
  % gallery set and A2 as probe set; Set-B and Set-C are probe sets corresponding to
  % carrying and clothing conditions respectively.  
\end{table}

The CASIA dataset B is the benchmark gait database used for the experimental
validation. The dataset includes six instances of normal walk (Set-A), two
instances of walking while carrying a bag (Set-B) and two instances of walking
while wearing an overcoat (Set-C) of 124 individuals. Each instance is captured
over 11 angles of view, from 0$\degree$ to 180$\degree$, adding up to a total of
13640 instances. Further detail can be obtained in \cite{yu2006framework}. Set-A
is split to Set-A1 containing four instances and Set-A2 for the remaining. Only
Set-A1 is used for training. 24 subjects were randomly selected from the CASIA-B
dataset to participate in the tuning set. These subjects were removed from the
gallery for the evaluation phase just as in \cite{jia2015view}.

The experiments were first executed under the sagittal angle, 90$\degree$ view,
to focus on the effect of carrying and clothing covariates. The GEI, GEnI, and
AEI were used as the base templates. The templates before and after GTS appear as
shown in Fig.~\ref{fig:gtsmask}. The performance of the proposed GTS is compared
against that claimed by other approaches in Table~\ref{tab:ccr}.

The upper portion of the gait template segmented by the GA chose only the head
of the subject and neglected the shoulders as opposed to what was selected by
Jia et al. in \cite{jia2015view}. The GA detected that the shoulder metric would
lead to a considerable loss in accuracy while wearing an overcoat and hence
chose $S_\text{H}$ a little before shoulder region. 

It is evident from the previously reported results in Table~\ref{tab:ccr} that
the clothing condition is the most challenging covariate leading to a lesser
CCR. Clothing conditions cause a greater change in the subjects' silhouettes. As
template-based methods rely on spatiotemporal changes of the silhouettes during
gait, the recognition performance is adversely affected. A more efficient
performance is attained when the regions that have an impact on such covariates
are masked out. The arm-swing constraints imposed by the weight of the clothing
and the carrying condition would compromise the accuracy at the midsection. As
speculated, the mid-left and mid-right sections were ignored in the optimal
hypothesis generated by the GTS for every angle and each type of gait
template. Note that the segmented GEI has a much smaller lower section due to
the greater effect of the covariates on the GEI template. The area permitted by
the mask is 25.2\% of the total template area; neglecting the constant features,
only 8.4\% of the feature space is utilized. Nevertheless, the GEI masked with
GTS outperforms the existing methods.

Genetic algorithm is known to have a tendency to give subobtimal results. There
comes a requirement to tune the parameters after the genetic algorithm
converges. The outcome of the GTS shows that only two parameters are variable:
$S_\text{H}$ and $S_\text{F}$. That is, weight bits are optimally assigned as
$[W_\text{H}, W_\text{L}, W_\text{R}, W_\text{F}]=[1,0,0,1]$. This assignment
leaves $S_\text{M}$ irrelevant as both mid-sections are ignored. These two
variables can be sequentially optimized starting with $S_\text{F}$ with a fixed
$S_\text{H}$ and then $S_\text{H}$ with the optimized $S_\text{F}$. This process
is also followed using the tuning set for validation.

\begin{table}
  \centering
  \caption{CCR(\%) Without Prior Knowledge of View Angle}
  \label{tab:ccr-angle}
  % RECOMPILE THIS TABLE BEFORE SUBMISSION
  % \includegraphics[width=\linewidth]{CCR-AllAngles.pdf}
  \resizebox{\linewidth}{!}{%
  \begin{tabular}{p{0.067\linewidth}*{12}{>{\raggedleft\arraybackslash}p{0.05\linewidth}}}
    \toprule
    Angle & $0\degree$ & $18\degree$ & $36\degree$ & $54\degree$ & $72\degree$ 
    & $90\degree$ & $108\degree$ & $126\degree$ & $144\degree$ & $162\degree$ & $180\degree$\\ 
 
    \cmidrule{2-12} 
    \rule{0em}{1em} &\multicolumn{11}{l}{(a) {Dupuis et al. 
                      \cite{dupuis2013feature} Panoramic Gait Recognition on GEI}}\\
    \mbox{Normal} & 97.17 & 99.60 & 97.15 & 96.33 & 98.76 & 98.43 
                  & 97.14 & 97.57 & 97.14 & 92.97 & 96.00\\
    \mbox{Bag} & 73.15 & 74.07 & 74.70 & 76.33 & 78.49 & 75.81 
                  & 76.29 & 76.71 & 73.41 & 73.19 & 74.56\\
    \mbox{Coat} & 81.64 & 87.39 & 86.29 & 84.34 & 89.96 & 91.86 
                  & 89.50 & 85.04 & 72.24 & 78.40 & 82.70\\
    Mean & 83.99 & 87.02 & 86.05 & 85.67 & 89.07 & 88.70 
                  & 87.64 & 86.44 & 80.93 & 81.52 & 84.42\\
    \cmidrule{2-12} 
    \rule{0em}{1em} &\multicolumn{11}{l}{(b) {Choudhury et al. 
                      \cite{choudhury2015robust} View-Invariant Multiscale Gait Recognition on GEI}}\\
    \mbox{Normal} & 100.0 & 99.00 & 100.0 & 99.00 & 100.0 & 100.0 
                  & 99.00 & 99.00 & 100.0 & 100.0 & 99.00\\
    \mbox{Bag} & 93.00 & 89.00 & 89.00 & 90.00 & 77.00 & 80.00 
                  & 82.00 & 84.00 & 92.00 & 93.00 & 89.00\\
    \mbox{Coat} & 67.00 & 56.00 & 80.00 & 71.00 & 75.00 & 77.00 
                  & 75.00 & 65.00 & 64.00 & 64.00 & 66.00\\
    Mean & 86.67 & 81.33 & 89.67 & 86.67 & 84.00 & 85.67 
                  & 85.33 & 82.67 & 85.33 & 85.67 & 84.67\\
    \cmidrule{2-12} 
    \rule{0em}{1em} &\multicolumn{11}{l}{(c) {Rida et al. 
                      \cite{rida2016human} Group Lasso of Motion on GEI}}\\
    \rule{0em}{1.2em}%
    \mbox{Normal} & 97.97 & 98.79 & 96.37 & 96.77 & 98.39 & 97.98 
                  & 97.18 & 95.56 & 96.77 & 97.98 & 97.58\\
    \mbox{Bag} & 72.76 & 72.58 & 75.81 & 76.42 & 75.81 & 73.66 
                  & 74.60 & 76.92 & 76.11 & 75.10 & 76.11\\
    \mbox{Coat} & 80.49 & 83.47 & 85.08 & 87.85 & 91.53 & 91.07 
                  & 87.90 & 86.23 & 87.45 & 84.90 & 83.06\\
    Mean & 83.74 & 84.95 & 85.75 & 87.01 & 88.58 & 87.57 
                  & 86.56 & 86.24 & 86.78 & 85.99 & 85.58\\
    \cmidrule{2-12} 
    \rule{0em}{1em} &\multicolumn{11}{l}{(d) {Proposed 
                      Genetic Template Segmentation on GEI}}\\
    \rule{0em}{1.2em}%
    \mbox{Normal} & 98.50 & 98.98 & 99.00 & 97.00 & 97.50 & 96.00 
                  & 95.00 & 97.50 & 94.00 & 93.85 & 98.99\\
    \mbox{Bag} & 95.00 & 98.47 & 96.50 & 96.00 & 97.50 & 93.50 
                  & 93.50 & 94.00 & 92.50 & 91.33 & 94.44\\
    \mbox{Coat} & 97.00 & 99.49 & 97.50 & 94.00 & 88.00 & 90.50 
                  & 89.50 & 94.50 & 92.00 & 91.28 & 93.94\\
    Mean & \textbf{96.83} & \textbf{98.98} & \textbf{97.67} & \textbf{95.67} 
         & \textbf{94.33} & \textbf{93.33} & \textbf{92.67} & \textbf{95.33} 
         & \textbf{92.83} & \textbf{92.15} & \textbf{95.79}\\
    \bottomrule
  \end{tabular}%
  }
\end{table}

The GTS is applied so as to generate one masking template for every angle using
the tuning set. The tuning set is also used to train the view estimator. The
evaluation set is separated into gallery and probe sets. After which, 11
LDA-Bayes' classifiers are trained (one for each view angle) using the gallery
set. The angle of each instance of the probe set is predicted with
the view estimator. The instance is then passed to the appropriate view-specific
classifier for the identity prediction. Note that each angle set also has it's own
PCA-LDA transformation. PCA is set to retain $99\%$ of data variance. This resulted in retaining a different number of Eigenvectors for each angle for a given template. The numbers range from 123 to 181 for GEI, 147 to 181 for GEnI, and 95 to 147 for AEI.

\begin{table}
  \centering
  \caption{View-invariant CCR Comparison}
  \label{tab:ccr-viewinvar}
\resizebox{\columnwidth}{!}{%
  \begin{tabular}{p{2.5cm} *{5}{p{0pt} c}}
    \toprule
    Method && Normal && Bag && Coat && Mean && Std \\
    \cmidrule(lr){1-1}
    \cmidrule(lr){3-7}
    \cmidrule(lr){9-11}
    GEI with PGR  \cite{dupuis2013feature} && 97.11 && 75.16 && 84.49 && 85.59 && 11.02\\
    \mbox{GEI with VI-MGR~\cite{choudhury2015robust}} && \textbf{99.55} 
                     && 87.09 && 69.09 && 85.24 && 15.31\\
    GEI with GLM \cite{rida2016human} && 97.39 && 75.08 && 86.28 && 86.25 && 11.16\\
    Whole GEI && 98.12 && 81.77 && 32.66 && 70.85 && 34.07\\
    GEI with GTS && 96.94 && \textbf{94.79} && \textbf{93.43} && \textbf{95.05} && \textbf{1.77}\\
    Whole GEnI && 96.76 && 84.41 && 40.64 && 73.94 && 29.49\\
    GEnI with GTS && 95.11 && 92.52 && 91.32 && 92.98 && 1.94\\
    Whole AEI && 95.62 && 75.51 && 42.42 && 71.18 && 26.86\\
    AEI with GTS && 90.61 && 85.58 && 77.71 && 84.63 && 6.50\\
    \bottomrule
  \end{tabular}%
}
\end{table}

\begin{figure}
  \centering
  \includegraphics[width=0.95\linewidth]{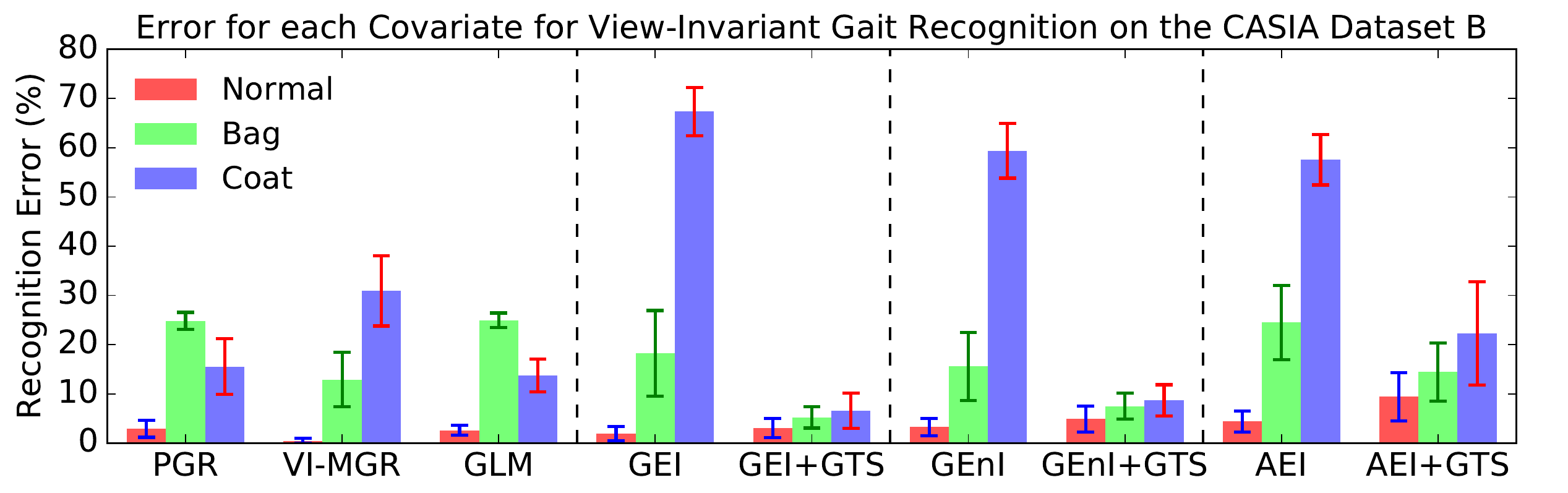}
  \caption{Recognition error %for each covariate with different algorithms and various templates
    without prior knowledge of view angle. Error lines depict
    the standard deviation of the error taken over the 11 views.}
  \label{fig:covarerror}
\end{figure}

The accuracy of the view estimator plays a vital role in view-invariant
recognition. The proposed view estimator is $97.77\%\pm 1.57$ accurate in
finding the correct angle of the given gait sequences in contrast to the
$94.43\%\pm 1.39$ proposed in \cite{dupuis2013feature}. In addition, the
view-dependent classifiers are also capable of producing an applicable accuracy
to neighboring views minimizing the error of the overall recognition. 

Table~\ref{tab:ccr-angle} reports the CCR of the state-of-the-art view-invariant
gait recognition methods along with the best performing template with the GTS,
the GEI. All of the scores in this table have been claimed to be obtained
without the prior knowledge of the actual view angle. The overall performance of
the methods including the base templates taking into account all angles is
provided in Table~\ref{tab:ccr-viewinvar}. Fig.~\ref{fig:covarerror} compares the
error associated for each covariate for different methods. It is evident that
the GTS has improved the covariate performance of all of the base gait templates.

The VI-MGR shows the highest normal condition CCR, but with a substantially
lower CCR for the clothing condition. The PGR and GLM perform equally well with
a slight trade-off in carrying condition. The GTS with the GEI shows the best
CCR in both carrying and clothing condition with minimal trade-off in normal
condition resulting in a far superior overall performance. The
  entire operation was also implemented with $k$NN in place of Bayes' rule for
  comparison. On an average of all 11 views and 3 covariates, GTS-GEI with $k$NN
  (k=$1$) yeilded an accuracy of 94.54\% which is marginally lesser than
  Bayes' rule with 95.05\%.

\section{Conclusion and Future Work}
In this paper, a novel segmentation technique was proposed to find the optimal
regions of a gait template for view-invariant gait recognition robust to
covariate factors. The genetic algorithm automates the boundary selection for
each angle while a view-estimator determines the probe angle and selects the
suitable view-specific classifier for recognition. The overall results clearly
depict that the proposed GTS method outperforms the existing methods in
literature. The next step would be to extend this framework to gait
authentication.

%\section*{Acknowledgment}

\newpage
\ifCLASSOPTIONcaptionsoff
  \newpage
\fi

\IEEEtriggeratref{10}


\IEEEtriggercmd{\enlargethispage{-2.5in}}

\bibliographystyle{IEEEtran}
%\bibliography{GR-GTS}

\end{document}